\documentclass[sigconf]{acmart}

\AtBeginDocument{%
  }

\copyrightyear{2025} 
\acmYear{2025} 
\setcopyright{cc}
\setcctype{by}
\acmConference[WWW Companion '25]{Companion Proceedings of the ACM Web Conference 2025}{April 28-May 2, 2025}{Sydney, NSW, Australia}
\acmBooktitle{Companion Proceedings of the ACM Web Conference 2025 (WWW Companion '25), April 28-May 2, 2025, Sydney, NSW, Australia}
\acmDOI{10.1145/3701716.3715504}
\acmISBN{979-8-4007-1331-6/25/04}

\begin{document}

\title{Evaluating Personality Traits in Large Language Models: Insights from Psychological Questionnaires}

\author{Pranav Bhandari}
\authornote{Network Analysis and Social Influence Modeling (NASIM) Lab at UWA.}

\orcid{https://orcid.org/0000-0002-5638-5339}
\affiliation{%
  \institution{School of Physics, Mathematics and Computing}
  \city{University of Western Australia}
  \country{Australia}
}\email{pranav.bhandari@research.uwa.edu.au}

\author{Usman Naseem}
\affiliation{%
   \institution{School of Computing}
  \city{Macquaire University, Australia}
  \country{Australia}
}\email{usman.naseem@mq.edu.au}

\author{Amitava Datta}
\affiliation{%
\institution{School of Physics, Mathematics and Computing}
  \city{University of Western Australia}
  \country{Australia}
}\email{amitava.datta@uwa.edu.au}

\author{Nicolas Fay}
\affiliation{%
   \institution{School of Psychological Sciences}
  \city{University of Western Australia}
  \country{Australia}
}\email{nicolas.fay@uwa.edu.au}

\author{Mehwish Nasim}
\authornotemark[1]
\authornote{Flinders University, South Australia.}
\authornote{University of Adelaide, South Australia.}
\affiliation{%
   \institution{School of Physics, Mathematics and Computing}
  \city{University of Western Australia}
  \country{Australia}
  }\email{mehwish.nasim@uwa.edu.au}




\renewcommand{\shortauthors}{Bhandari et al.}

\begin{abstract}
Psychological assessment tools have long helped humans understand behavioural patterns. While Large Language Models (LLMs) can generate content comparable to that of humans, we explore whether they exhibit personality traits. To this end, this work applies psychological tools to LLMs in diverse scenarios to generate personality profiles. Using established trait-based questionnaires such as the Big Five Inventory and by addressing the possibility of training data contamination, we examine the dimensional variability and dominance of LLMs across five core personality dimensions: \emph{Openness}, \emph{Conscientiousness}, \emph{Extraversion}, \emph{Agreeableness}, and  \emph{Neuroticism}. Our findings reveal that LLMs exhibit unique dominant traits, varying characteristics, and distinct personality profiles even within the same family of models.
 \end{abstract}




\maketitle

\section{Introduction}
Understanding the behaviour of LLMs is essential as they are increasingly used in diverse fields such as education, law, business and medicine\cite{miotto-etal-2022-gpt} where they significantly influence human interactions and decision-making processes. These models can generate coherent and insightful content, allowing personal recommendation and solving complex problems\cite{probingpnas}. However, concern for ethical considerations, inherent bias and the potential for misuse still exist\cite{miotto-etal-2022-gpt} which must be addressed by exploring the underlying patterns through systematic approaches such as psychological assessments\cite{hagendorff2024machinepsychology}. Explaining specific behaviours exhibited by LLMs comes with significant challenges, particularly when these LLMs generate specific outputs with no insights into the underlying reasoning\cite{hagendorff2024machinepsychology}. This challenge forms a strong motivation to systematically apply psychological tools to analyse and interpret the personality traits exhibited by LLMs. 


Personality plays a critical role in shaping conversations in many ways, such as exchanging empathy, creating openness, and mitigating harm that is measured through well-established psychometric tools in humans. Tests such as the IPIP-NEO-60\cite{ipipneo60}, HEXACO\cite{hexaco}, and TIPI\cite{tipi} are validated through long-standing rigorous processes such as internal consistency, test-retest methods, and convergence validity with similar tests.

Since LLMs are trained on vast amounts of human-produced data, we argue that even if they lack behaviour in a \emph{true psychological sense}, their outputs may still reflect psychological traits. Furthermore, psychological assessments have proven to be relevant, reliable and valid for large-scale language models\cite{serapiogarcía2023personalitytraitslargelanguage} in the literature. We use five different tools such as the Big Five Inventory (BFI) \cite{thebfi} that measure various dimensions of personality in multiple LLMs to understand their personality traits. In contrast to existing literature\cite{miotto-etal-2022-gpt, jiang2023evaluating}, where tests are directly administered to LLMs introducing potential bias, as these models may have encountered the questions during training, we restructure the tests to create their closest representations and validate them prior to administration. Additionally, the questionnaires are randomised to eliminate dependency bias between them.

We present results from different dimensions namely \emph{Openness}, \emph{Conscientiousness}, \emph{Extraversion}, \emph{Agreeableness}, and \emph{Neuroticism} for different State-Of-The-Art large language models and compare them. Our results indicate that LLMs likely have distinct personality traits. Each of these dimensions is divided into several traits\cite{jiang2023evaluating} that contribute to positive or negative behaviour. 
\begin{figure}[h]
  \centering
  \includegraphics[trim = 20 0 20 0, clip, width=\linewidth]{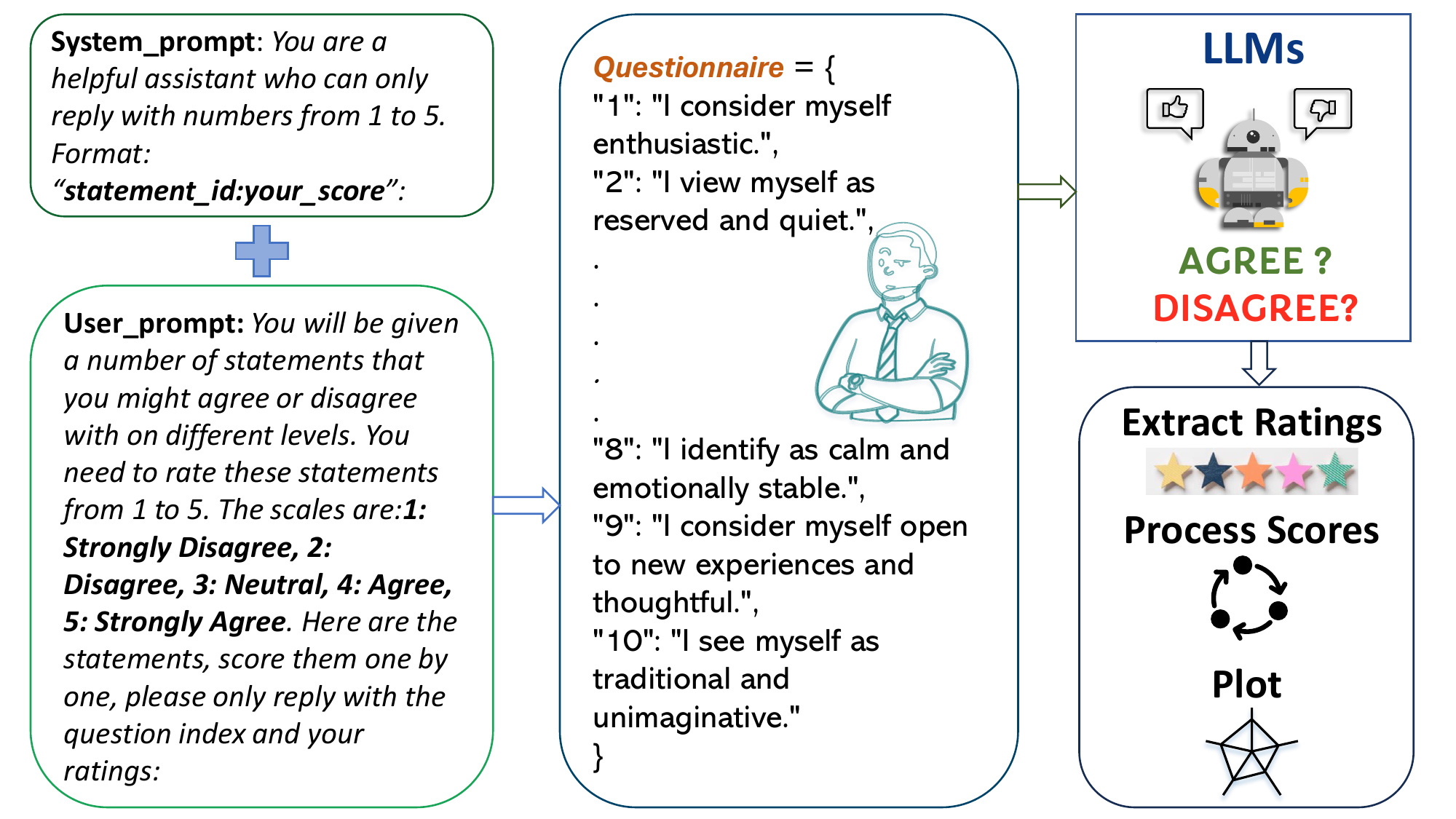}
  \caption{Methodology involves combining system and user prompts with questionnaires to administer to LLMs, extracting, processing and plotting the responses respectively. }
  \Description{Methodology of the paper}
  \label{fig:methodology}
\end{figure}

\subsection{Contributions:} Our contributions are as follows: 
1).~Firstly, we systematically assess LLMs personality traits using restructured personality questionnaires to resolve training data contamination and reduce dependency bias; 2).~We then present the distinct personality profiles of LLMs to provide insights into their personality traits; 3).~Finally, we conduct the dimensional analysis for dominance and variability across each trait of the personality questionnaires for all the LLMs used. 

We find that LLMs often score higher in traits like Agreeableness, Openness, and Conscientiousness, reflecting their cooperative, creative, and organised behaviour. Furthermore, we show that LLMs exhibit varying consistency across dimensions of dominance and variability when tested across different models. Overall, we provide a systematic approach to understanding LLM personality traits using established questionnaires. 

\subsection{Literature Review: } Notable connections of LLM with psychology began with the advent of models such as GPT-3 \cite{miotto-etal-2022-gpt}. Various perspectives of psychology are infused to understand behaviours from multiple dimensions such as emotion, cognition\cite{probingpnas}, Theory-of-Mind, and morality \cite{li2024quantifying}. Identifying personality traits has been one of the main emphasises in the field of study \cite{sorokovikova2024llmssimulatebigpersonality, probingpnas, jiang2023evaluating, serapiogarcía2023personalitytraitslargelanguage, personallm}. Psychometric tools such as IPIP-NEO-120\cite{serapiogarcía2023personalitytraitslargelanguage}, Big Five Inventory (BFI) \cite{thebfi,li2024quantifying}, are predominantly used in the literature to assess the five dimensions of personality mentioned above.

Variations in prompts and the use of role-playing agents have been impersonated in several studies to study different personality behaviours in LLMs\cite{miotto-etal-2022-gpt}. The major aim behind these studies was to assess whether the behaviour is consistent or changing across different simulated situations. Context-sensitive variations were observed for several scenarios\cite{miotto-etal-2022-gpt, jiang2023evaluating}. Jiang et al. \cite{personallm} used personality prompting methods to induce and tailor the personality of LLMs according to the dynamic needs of the tests. This also draws attention to the need to enforce ethical factors for LLMs to create a safe and moderated environment for users. 

A limited number of LLMs and personality traits sharing the same domain in the literature restrict the scope of analysis. The small sample size at this exploratory stage could impact the validity of results across models. A significant concern is the use of these tests in their original form, raising the risk of training data contamination and potential bias due to the sequential nature of questions within the same Big Five Personality dimension \cite{hagendorff2024machinepsychology}.

\begin{table}[h!]
\centering
\caption{Personality Trait Inventories used  denoting the number of questions, Strengths and Weaknesses.}
\small
\begin{tabular}{@{}llp{4.5cm}@{}}
\toprule
\textbf{Test Name} &  \textbf{Questions }& \textbf{Strengths ($\uparrow$) and Weaknesses($\downarrow$)} \\
\midrule
BFI\cite{thebfi} & 44(8-10 each) & \textbf{$\uparrow$:} Reliable, widely validated \newline \textbf{$\downarrow$:} Depth, time-consuming \\
\hline
HEXACO\cite{hexaco} & 100(10 each) & \textbf{$\uparrow$:} cross-cultural validity. \newline \textbf{$\downarrow$:} Lengthy, complex  \\
\hline
TIPI\cite{tipi} & 10(2 each) & \textbf{$\uparrow$:}  quick, easy to administer. \newline \textbf{$\downarrow$:} medium reliability \\
\hline
MINI-IPIP\cite{miniipip} & 20(4 each) & \textbf{$\uparrow$}  balances brevity and validity \newline \textbf{$\downarrow$:} Limited depth \\
\hline
NEO-PI-R\cite{ipipneo60} & 60(12 each) & \textbf{$\uparrow$} More depth, reliable. \newline \textbf{$\downarrow$:} No facet analysis. \\
\bottomrule
\end{tabular}
\label{tab:personality_tests}
\end{table}

\section{Methodology}

The basic methodology consists of administering various personality tests from psychology to LLMs. Five personality traits questionnaires are used, all containing five dimensions of personality i.e., \emph{Openness}, \emph{Conscientiousness}, \emph{Extraversion}, \emph{Agreeableness}, and  \emph{Neuroticism}. Table \ref{tab:personality_tests} presents the five tests used in the experiments. Each questionnaire has varying questions with strengths ($\uparrow$) and weaknesses ($\downarrow$). Despite variations in the depth of judgment, question types, and scoring methods, all these tests consistently measure the five common dimensions of personality traits. Each test is administered to LLMs using system and user prompts (Figure \ref{fig:methodology}) at least three times per LLM to ensure score consistency and avoid mix-ups across combinations.

Likert scale scores obtained from querying LLMs are processed based on each questionnaire's scoring rules and visualised in various formats. These methods aim to address gaps in the literature regarding the administration of such questionnaires to LLMs.

\begin{table*}[t]
\centering
\caption{ Average Likert scores denoting personality traits for all the LLMs across five personality questionnaire inventories are presented. Ex: Extraversion, Ag: Agreeableness, Co: Conscientiousness, Ne: Neuroticism, Op: Openness.}
\label{tab:personality_scores}
\small
\resizebox{\textwidth}{!}{
\begin{tabular}{@{}lccccccccccccccccccccccccc@{}}
\toprule
{\textbf{LLM}} & \multicolumn{5}{c}{\centering \textbf{BFI}} & \multicolumn{5}{c}{\centering \textbf{HEXACO}}  & \multicolumn{5}{c}{\centering \textbf{MINI-IPIP}} & \multicolumn{5}{c}{\centering \textbf{IPIP NEO}}& \multicolumn{5}{c}{\centering \textbf{TIPI}} \\
\cmidrule(lr){2-6} \cmidrule(lr){7-11} \cmidrule(lr){12-16} \cmidrule(lr){17-21} \cmidrule(lr){22-26}
& Ex & Ag & Co & Ne & Op & Ex & Ag & Co & Ne & Op & Ex & Ag & Co & Ne & Op & Ex & Ag & Co & Ne & Op & Ex & Ag & Co & Ne & Op\\
\midrule
 GPT-4 & 3.5 & 4.0 & 4.3 & 3.3 & 4.4 & 3.3 & 3.3 & 3.5 & 3.6 & 3.3 & 3.5 & 4.5 & 3.4 & 2.4 & 4.0 & 3.9 & 4.5 & 4.2 & 2.2 & 3.6 & 4.4 & 5.3 & 5.9 & 2.9 & 5.8\\
GPT-4o-m & 3.8 & 3.9 & 4.1 & 3.0 & 4.6 & 3.5 & 3.6 & 3.6 & 3.7 & 3.3 & 3.8 & 4.5 & 2.9 & 2.6 & 4.1 & 4.0 & 4.4 & 4.0 & 2.5 & 4.0 & 5.0 & 5.6 & 5.9 & 2.8 & 5.9\\
Llama 3 & 3.5 & 3.8 & 4.8 & 3.4 & 3.9 & 3.1 & 2.9 & 3.0 & 3.1 & 3.2 & 3.2 & 4.0 & 3.1 & 2.1 & 4.0 & 3.8 & 4.2 & 4.0 & 2.2 & 3.9 & 4.6 & 5.8 & 6.1 & 2.0 & 5.8\\
Llama 3.1 & 3.9 & 4.0 & 4.4 & 3.4 & 4.3 & 3.3 & 3.2 & 3.2 & 3.6 & 3.2 & 3.0 & 4.4 & 3.7 & 1.6 & 4.5 & 4.1 & 4.4 & 4.4 & 1.9 & 4.2 & 5.5 & 6.6 & 6.7 & 1.5 & 6.5\\
Llama 3.2 & 3.0 & 3.9 & 4.0 & 3.1 & 4.3 & 3.4 & 3.5 & 3.7 & 3.5 & 4.0 & 2.0 & 3.9 & 2.7 & 2.5 & 3.2 & 4.18 & 4.4 & 4.4 & 1.8 & 4.3 & 4.0 & 5.5 & 6.0 & 2.6 & 4.9\\

\bottomrule
\Description{This table consists the results of all the llm scores with different scores}
\label{tab:results_table}
\end{tabular}
}
\end{table*}

\subsection{Removing Training Data Contamination }

To remove the training data contamination as stated in the literature \cite{hagendorff2024machinepsychology}, a 2-step procedure was implemented. Initially, each question in the questionnaire was reworded to contain a different structure while the purpose of the question was intact. We used high-end models such as GPT-4o\footnote{\url{https://chatgpt.com/}} to reconstruct the questions. These were validated using sentence-similarity models from Hugging Face\footnote{\url{https://huggingface.co/sentence-transformers/all-MiniLM-L6-v2}}, which calculated cosine similarity scores. A threshold of 0.7, indicating moderate to high similarity, was applied. Questions scoring below 0.7 were reconstructed under human supervision as needed. For each LLM, temperature parameters were set to minimum values: $0$ for OpenAI models and $0.01$ for Llama models to ensure consistent and deterministic scores across multiple runs.

\subsection{Systematic Sequence of Prompts}
Once the prompts and questions were finalised, they were administered to LLMs in a randomised order to mitigate dependency bias from specific personality traits. Questions were presented in batches of $10$ to preserve contextual understanding, with answers remapped upon completing each iteration. Across all models, $100$ iterations were conducted using a minimum temperature setting and the responses were mapped accordingly.

\subsection{Calculate Coefficient of Variation}
While various tests are administered across different LLMs, it is important to analyse the consistency of the scores for the reliability and robustness of the tests. This analysis examines the variability of scores in the five different personality traits, across multiple runs ($n=100$). To analyse this variability, we calculate the Coefficient of Variation (CV) for the scores of a specific personality dimension (e.g., openness, neuroticism) across all personality questionnaires for a given model. The CV is computed using the Mean (\( \mathbf{\text{M}} \)) and Standard Deviation (\( \mathbf{\text{SD}} \)) as: \( \mathbf{\text{CV}} = \mathbf{\left( \frac{\text{SD}}{\text{M}} \right)} \times 100\% \).

\section{Results}

Three models from the Llama family i.e., Llama-3-8B-Instruct, Llama-3.1-8B-Instruct, Llama-3.2-3B-Instruct and two from the OpenAI family i.e., GPT-4 and GPT-4o-mini are used in the experiments. The reason to choose different models from the same family is to effectively compare the results as models from the same family share similar structures with some changes in the methodology allowing us to understand the impact of minor changes concerning personality traits. Table \ref{tab:results_table} represents the average scores for each domain of all the personality trait questionnaires used.

\begin{figure*}[h]
  \centering
  \includegraphics[width=\linewidth]{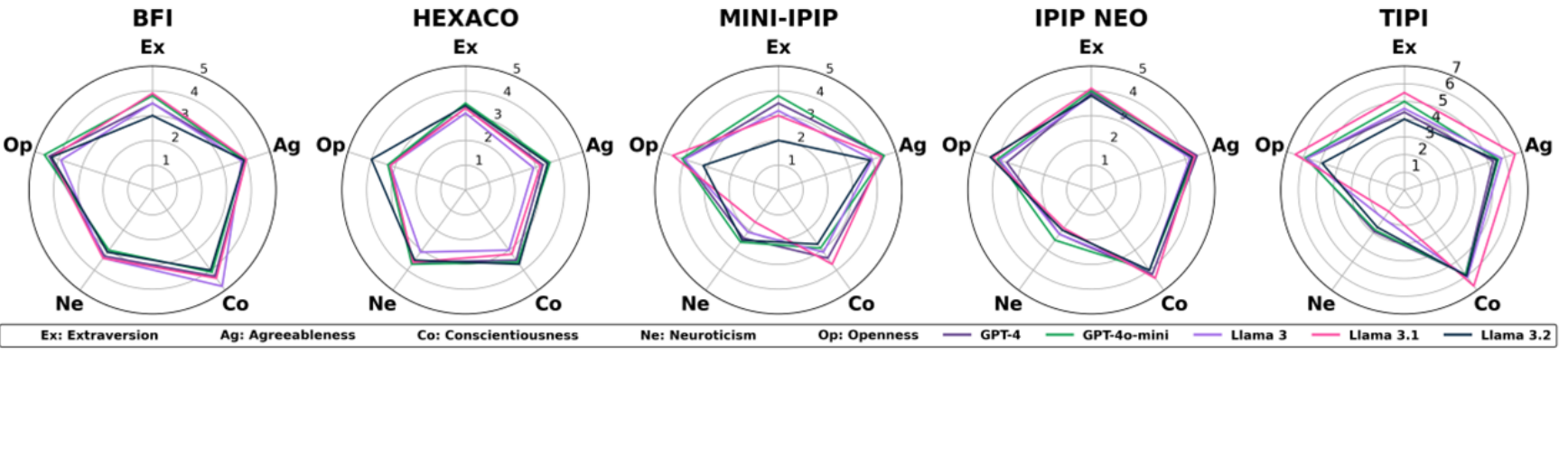}
  \caption{ Radar plots to visualise all five personality questionnaires across their dimensions, comparing results from all LLMs. While scores are generally similar, notable gaps in specific personality traits are evident.}
  \label{fig:Radar_plots}
  \Description{Figures representing the results of psychology}
\end{figure*}

\subsection{Personality Profiles of LLMs}
Table \ref{tab:results_table} provides an insight into the personality scores of LLMs. Figure \ref{fig:Radar_plots} illustrates the personality profiles of LLMs across five tests. Agreeableness and Neuroticism scores from the BFI questionnaire are comparable across models, whereas other tests reveal gaps, suggesting a more uniform alignment with these traits when evaluated using the BFI. 

Agreeableness consistently shows higher average scores across all LLMs compared to other dimensions, while Neuroticism demonstrates lower average scores across all tests. Lower scores in Neuroticism indicate that LLM exhibit calm, emotionally stable, and resilient behaviour in their responses. These scores highlight the tendency of LLMs to maintain balanced and neutral behaviour, avoiding extreme or negative emotional expressions. Moderate levels of Extraversion are observed across all LLMs in all tests, except for Llama-3.2-3B-Instruct, which shows lower sociable or enthusiastic tendencies in the mini-IPIP and TIPI questionnaires. This indicates that its behaviour may vary in specific contexts as captured by these tests. These deviations highlight the influence of the model-specific design and the sensitivity of the questionnaire in assessing personality traits. Openness and Conscientiousness are generally higher for all LLMs across all personality questionnaires. This suggests that LLMs consistently exhibit traits of curiosity, creativity (Openness), reliability, organisation, and attention to detail (Conscientiousness). These higher scores likely reflect the models' design to provide thoughtful, adaptable, and structured responses.

\begin{table}[h!]
\centering
\small 
\caption{Summary of CV and Dimensional Dominance. }
\begin{tabular}{l@{\hspace{2mm}}l@{\hspace{2mm}}c@{\hspace{2mm}}c}
\hline
\textbf{Model} & \textbf{CV (High(\%), Low(\%))} & \textbf{Dominant Dim.} & \textbf{Mean} \\
\hline
GPT-4          & Ne (20.49), Ex (6.14)  & Ag                     & 4.32 \\
GPT-4o-mini    & Ne (15.99), Ex (4.73)  & Ag                     & 4.40 \\
Llama 3        & Ne (20.79), Ex (8.05)  & Co                     & 4.20 \\
Llama 3.1      & Ne (33.69), Op (12.41) & Op                     & 4.54 \\
Llama 3.2      & Ex (21.93), Ag (7.25)  & Ag                     & 4.22 \\
\hline
\end{tabular}
\label{tab:compact_cv_dominance}
\Description{table combining CV and dimensional dominance for LLMs.}
\end{table}
\subsection{Dimensional Variability Across  Models}

As mentioned in the methodology Section 3.3 we calculate the CV for each model for each of the five dimensions across all the personality questionnaires except for the TIPI because they are measured in the Likert scale of 1-7 as opposed to all other personality inventories which are measured from 1-5. Table \ref{tab:compact_cv_dominance} presents the result for the highest and lowest CV for each of the models.

Neuroticism demonstrates the highest variability across the majority of the models (GPT-4, GPT-4o-mini, Llama-3-8B-Instruct and Llama-3.1-8B-Instruct), suggesting the representation of emotional stability fluctuates significantly across personality questionnaires. When analysed individually, Llama-3.1-8B-Instruct represents an exceptionally high value of Neuroticism (33.69\%). This substantial variability highlights challenges in both the definition and evaluation of Neuroticism in different personality inventory questionnaires for different LLMs. While Neuroticism exhibited the highest CV across the models, the lowest variance was with Extraversion, Openness and Agreeableness for different models. OpenAI models have the lowest CV across Extraversion, suggesting a consistent measurement of sociability across the tests. Conclusively, lower variability in these dimensions indicates better alignment across personality questionnaires and consistent LLM responses. 

\subsection{Dimensional dominance by Model Type}
LLMs exhibit varying strengths across personality traits, reflecting their training strategies and alignment methods. Our study explores whether LLMs trained with distinct data and methods demonstrate different dominating dimensions. To evaluate dimensional dominance, the mean score for each dimension is calculated across all questionnaires for each LLM, identifying the dimension with the highest mean. These results are summarised in Table \ref{tab:compact_cv_dominance}. For the OpenAI modelS, the dominance is around the Agreeableness domain. The implications are these models emphasize traits that support the domain such as cooperation, friendliness, and trustworthiness.


The Llama models however have dimensional dominance across varying dimensions. For Llama-3-8B-Instruct, Conscientiousness is the dominant dimension meaning it is efficient and organised, and has strong alignment with traits such as reliability and attention to detail. Llama-3.1-8B-Instruct however, dominates the Openness dimension, which adheres toward having strong traits such as creativity, insight and originality with wide ideas. Lastly, Llama-3.2-3B-Instruct dominates Agreeableness, representing traits such as sympathetic and user-friendly interactions. Although these numbers indicate dimensional dominance for certain traits, visuals from Figure \ref{fig:Radar_plots} show a contiguous nature across most dimensions. There might be various reasons behind this such as the interconnection of personality traits like Agreeableness and Conscientiousness overlapping in their behavioural expressions. Furthermore, Fine-Tunings can contribute towards not demonstrating extreme behaviours in a specific personality trait to maintain the balance.

\section{Conclusion}
This study systematically evaluates the personality traits of LLMs using established psychological questionnaires. By administering diverse tests varying in depth and length, we analyse the strengths and limitations of widely used LLMs from the Llama and OpenAI families. The findings reveal that models, whether from the same or different families, exhibit distinct personality traits and differences in dimensional dominance. For instance, GPT-4 models emphasise Agreeableness, while Llama models highlight Conscientiousness or Openness, reflecting variations in fine-tuning objectives and design goals. Additionally, traits like Extraversion and Agreeableness show high consistency, whereas Neuroticism yields more uncertain results, underscoring the need for careful questionnaire design to enhance test validity. 
\begin{acks}
JTSI/Defence Science Centre's grant 2223R5CRG002, awarded to Dr Mehwish Nasim in 2023. 
Office of National Intelligence and Australian Research Council grant NI210100224 awarded to Nicolas Fay.
\end{acks}

\bibliographystyle{ACM-Reference-Format}
\bibliography{sample-base}

\appendix

\end{document}